\documentclass[lettersize,journal]{IEEEtran}
\usepackage{amsmath,amsfonts}
\usepackage{algorithmic}
\usepackage{array}
\usepackage[caption=false,font=normalsize,labelfont=sf,textfont=sf]{subfig}
\usepackage{textcomp}
\usepackage{stfloats}
\usepackage{url}
\usepackage{verbatim}
\usepackage{graphicx}

\usepackage{amssymb}
\usepackage{booktabs}
\usepackage{balance}
\usepackage{makecell}
\usepackage{color}
\usepackage{multirow}

\def\BibTeX{{\rm B\kern-.05em{\sc i\kern-.025em b}\kern-.08em
    T\kern-.1667em\lower.7ex\hbox{E}\kern-.125emX}}
\usepackage{balance}

\begin{document}

\title{UBATrack: Spatio-Temporal State Space Model for General Multi-Modal Tracking}

\def\tracker{UBATrack}

\author{Qihua Liang, Liang Chen$^{\dagger}$, Yaozong Zheng$^{*}$, Jian Nong, Zhiyi Mo, Bineng Zhong$^{*}$
\thanks{Qihua Liang, Liang Chen, Yaozong Zheng, and Bineng Zhong are with the Key Laboratory of Education Blockchain and Intelligent Technology, Ministry of Education, Guangxi Normal University, Guilin 541004, China, the Guangxi Key Laboratory of Multi-Source Information Mining and Security, Guangxi Normal University, Guilin 541004, China, and University Engineering Research Center of Educational Intelligent Technology, Guangxi Normal University, Guilin, 541004, China.}
\thanks{Jian Nong and Zhiyi Mo is currently a Professor in the Guangxi Key Laboratory of Machine Vision and Intelligent Control, Wuzhou University, Wuzhou 543002, China.}
\thanks{Qihua Liang  and Liang Chen contributed equally. Yaozong Zheng and Bineng Zhong are corresponding authors.}
}

\markboth{Journal of \LaTeX\ Class Files,~Vol.~18, No.~9, September~2020}%
{How to Use the IEEEtran \LaTeX \ Templates}

\maketitle

\begin{abstract}
   Multi-modal object tracking has attracted considerable attention by integrating multiple complementary inputs (e.g., thermal, depth, and event data) to achieve outstanding performance.
   Although current general-purpose multi-modal trackers primarily unify various modal tracking tasks (i.e., RGB-Thermal infrared, RGB-Depth or RGB-Event tracking) through prompt learning, they still overlook the effective capture of spatio-temporal cues. In this work, we introduce a novel multi-modal tracking framework based on a mamba-style state space model, termed {\tracker}. Our {\tracker} comprises two simple yet effective modules: a Spatio-temporal Mamba Adapter (STMA) and a Dynamic Multi-modal Feature Mixer. The former leverages Mamba’s long-sequence modeling capability to jointly model cross-modal dependencies and spatio-temporal visual cues in an adapter-tuning manner. The latter further enhances multi-modal representation capacity across multiple feature dimensions to improve tracking robustness. In this way, {\tracker} eliminates the need for costly full-parameter fine-tuning, thereby improving the training efficiency of multi-modal tracking algorithms. Experiments show that {\tracker} outperforms state-of-the-art methods on RGB-T, RGB-D, and RGB-E tracking benchmarks, achieving outstanding results on the LasHeR, RGBT234, RGBT210, DepthTrack, VOT-RGBD22, and VisEvent datasets.

\end{abstract}


\section{Introduction}
\label{sec:intro}

\begin{figure}[t]
  \begin{center}
  \includegraphics[width=1\linewidth]{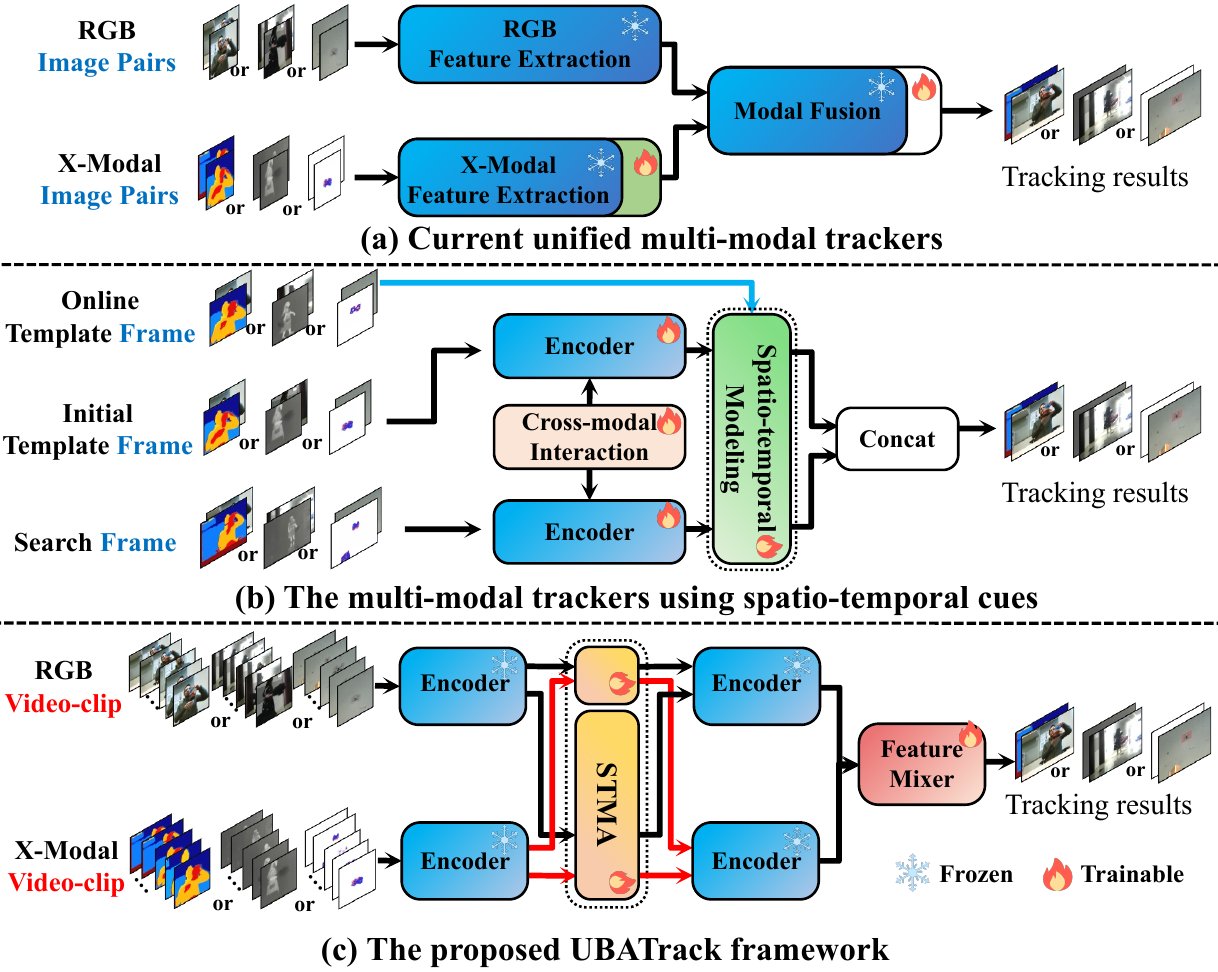}
  \end{center}
  \caption{
  (a) These trackers \cite{OneTracker,SDSTrack,un-track} primarily focus on modality fusion but often overlook effective spatio-temporal cue capture across modalities.
  (b) These trackers \cite{TATrack} learn temporal history information through full fine-tuning, which increases the training burden.
  (c) {\tracker} improves tracking performance by leveraging adapter-tuning to jointly model multi-modal spatio-temporal visual cues.
  }
  \label{fig:pic-01}
\end{figure}

\IEEEPARstart{V}isual object tracking \cite{SiamFC,SiamBAN,OSTrack} is fundamental in computer vision and is utilized in numerous vision tasks such as visual surveillance \cite{surveillance2}, video compression \cite{compression1}, and autonomous driving \cite{driving1}. It aims to continuously locate arbitrary objects in a video sequence relying solely on their initial positions. Traditional RGB tracking methods often suffer performance degradation in complex scenarios such as nighttime or occlusion, which limits their applicability in broader real-world settings. Consequently, an increasing number of researchers have begun to explore effective and low-cost ways of incorporating multi-modal cues into tracking frameworks, aiming to achieve more accurate target localization in challenging environments.

In recent years, many studies \cite{TBSI,BAT} have focused on modality-specific tracking tasks, attempting to explore effective multi-modal fusion strategies. For instance, TBSI \cite{TBSI} introduces a Template-Bridged Search region Interaction module to enable cross-modal interaction, thereby improving tracking robustness in challenging scenarios. However, the reliance on modality-specific designs limits their adaptability. More recently, to effectively aggregate multi-modal cues, several researchers \cite{ViPT,protrack} have proposed prompt learning multi-modal tracking frameworks, as illustrated in Fig.\ref{fig:pic-01}(a), which can independently handle multiple sub-tracking tasks without modifying the model architecture. For example, ViPT \cite{ViPT} adopts parameter-efficient tuning to learn a set of modality-relevant prompts tailored for different downstream tracking tasks. These approaches leverage prompt tuning to integrate complementary modality information into a pre-trained RGB tracking model, thus alleviating, to some extent, the scarcity of multi-modal training data. Furthermore, as shown in Fig.\ref{fig:pic-01}(b), a new dual-branch symmetric tracking framework \cite{OneTracker,SDSTrack,un-track} has been introduced, which fully fine-tunes multi-modal information while attempting to capture spatiotemporal contextual cues. 
UM-ODTrack \cite{UMODTrack}, for example, proposes two novel gated perceivers that adaptively learn cross-modal representations via a gated attention mechanism.
Although these efforts have led to significant improvements in tracking performance, such full fine-tuning paradigms impose substantial training overhead, making it difficult to efficiently exploit richer multi-modal spatio-temporal information.

To mitigate the aforementioned challenges, we explore state space models to efficiently capture spatio-temporal contextual knowledge, thereby enabling high-performance general multi-modal tracking. Specifically, we introduce a novel multi-modal tracking framework, termed {\tracker}, built upon a mamba-style state space model to unify diverse modality-specific tracking tasks, as shown in Fig.\ref{fig:pic-01}(c). Our {\tracker} comprises two simple yet effective modules: a Spatio-temporal Mamba Adapter (STMA) and a Dynamic Multi-modal Feature Mixer. The former leverages mamba’s long-sequence modeling capability to jointly model cross-modal dependencies and spatio-temporal visual cues in an adapter-tuning manner. The latter further enhances multi-modal representation capacity across multiple feature dimensions to improve tracking robustness. In this way, {\tracker} eliminates the need for costly full-parameter fine-tuning, thereby improving the training efficiency of multi-modal tracking algorithms. Extensive experiments demonstrate that {\tracker} outperforms state-of-the-art methods on RGB-T, RGB-D, and RGB-E tracking benchmarks, achieving outstanding results. 

    Our main contributions are summarized as follows:
    \begin{itemize}
        
    \item We introduce a new general multi-modal tracking framework based on state space models, which leverages adapter-based fine-tuning to efficiently unify multiple modality-specific tracking tasks and enhance the generalization of multi-modal tracking.

    \item We propose a Spatio-Temporal Mamba Adapter and a Dynamic Multi-modal Feature Mixer. The former is designed to capture spatio-temporal dependencies across diverse modalities, while the latter further strengthens the unified multimodal representation to enhance the robustness of the model.
    
    \item Extensive experiments on six benchmarks with various modality, including LasHeR \cite{LasHeR}, RGBT234 \cite{RGBT234}, RGBT210 \cite{RGBT210}, DepthTrack \cite{DeT}, VOT-RGBD22 \cite{DMTrack}, and VisEvent \cite{VisEvent}, show that {\tracker} achieves state-of-the-art performance and robust tracking.
    
    \end{itemize}   

\section{Related Work}
\label{sec:Related Work}

With the advancement of deep learning techniques \cite{He2025Unified,He2025Location,He2025Modular,He2025Waste,li2024towards,yin2025knowledge,dai2025unbiased,tcsvt1,gao2025knowledge,gao,gao25,FineCIR,HABIT,ReTrack,ENCODER,gong2024cross,gong2021eliminate,gong2022person,peng2025boosting,peng2025directing,peng2024lightweight,he2023degradation,he2023strategic,he2023camouflaged,lu2024robust,lu2023tf,lu2024mace,zhang2024cf,zhang2023multi,wang2024computing,ni2025recondreamer,ni2025wonderturbo,zeng2025FSDrive,zeng2025janusvln}, visual object tracking has attracted increasing attention from the research community.
According to various data modalities, visual tracking is primarily categorized into sub-learning tasks, including RGB tracking and multi-modal tracking (i.e., RGB-T/D/E tracking).

\subsection{Multi-modal object tracking}

    Visual object tracking aims to locate the coordinates of an object in a video sequence based on its initial positions. It includes RGB tracking \cite{SiamPIN,OSTrack,chen2025improving,SSTrack,li2025robust} and multi-modal tracking \cite{MMTrack,TBSI,ViPT}. RGB trackers only utilize raw RGB images for object tracking. This makes the trackers highly challenging in complex scenarios such as occlusion, low illumination, and fast movement, and they are prone to failures.
    This limitation prompts research into adding additional modalities to enhance performance. For example, depth cues improve the tracking ability to locate objects at different camera distances \cite{DeT,rgbd-0,gmmt}. Thermal camera data helps the trackers \cite{TBSI,BAT,TATrack} deal with challenges such as insufficient illumination, while event cameras improve the temporal perception level of the trackers \cite{VisEvent,event-0}. 
    In parallel, a robust multi-source framework\cite{ke2025early} was established by synergizing blockchain metrics, social sentiment, and regulatory signals, demonstrating that effectively capturing multi-modal dependencies can significantly improve predictive power in highly uncertain environments.
    These trackers are limited to specific tasks and cannot adapt to multiple tasks simultaneously with a single training process. Inspired by prompt learning \cite{VPT}, some works \cite{ViPT,OneTracker,SDSTrack,un-track} achieve general tracking for multiple tasks. However, they overlook multi-modal spatio-temporal context modeling, leading to weak performance in handling target appearance variations. Recently, some single-task multi-modal trackers \cite{TATrack, SPT, event-0} use historical modality frames and traditional score-based template updates. This weakens the discriminative power due to poor updates and requires extra modules for spatio-temporal modeling, adding complexity.

\subsection{Vision Mamba}

    Currently, Vision Transformers \cite{ViT} are the main backbone network due to their excellent performance in computer vision. But the quadratic complexity of attention causes huge computational overhead in downstream tasks with large spatial resolutions, deterring many researchers.
    To further reduce the complexity, researchers propose the State Space Model (SSM) \cite{ssm1,ssm2}. Building on this, Gu et al. \cite{mamba} propose Mamba, a new architecture with linear complexity and long-sequence modeling ability. Compared to Transformers, mamba technology \cite{simba} achieves similar performance at a lower computational cost and better preserves long-range dependencies without needing to store and process cache for K and V values. It has been widely explored in visual tasks, including detection \cite{mambaDetection1}, segmentation \cite{yang2025uncertainty}, and video understanding \cite{videomamba}, among others. 
    Inspired by the capability of Mamba, we design a Spatio-Temporal Mamba Adapter for {\tracker} to efficiently learn spatio-temporal representations of target instance across diverse modalities, which helps enhance the tracker’s robustness in complex scenarios.

\subsection{Vision MLP models}

    Recently, MLP-based vision models \cite{mlp-like1,mlp-like2} are proposed to reduce the inductive bias further and show competitive performances with CNN-based models and ViT \cite{ViT} on mainstream visual recognition tasks.
    S$^{2}$-MLP \cite{S2-mlp} uses spatial shifting and channel mixing for token interaction.
    Cycle-MLP \cite{cyclemlp} samples local points cyclically along the channel dimension for a more informative receptive field. 
    Hire-MLP \cite{Hire-mlp} rearranges regions to mix tokens and enable cross-region communication. 
    Strip-MLP \cite{Strip-MLP} uses cascaded group and local strip mixing modules to enhance token interaction by aggregating adjacent tokens across strips.
    Most of these methods use static fusion, focusing on spatial locality over high-level semantics. To address this, DynaMixer \cite{Dynamixer} creates dynamic mixing matrices from token content and improves visual recognition with dimensionality reduction and multi-segment fusion.
    Current general-purpose multi-modal trackers lack dynamic cross-modal fusion methods and fail to effectively learn discriminative features between modalities during this process.
    Inspired by the above approaches, we further design a Dynamic Multi-modal Feature Mixer to adaptively enhance the discriminative capability of multi-modal features for final target localization.

\begin{figure*}
    \begin{center}
    \includegraphics[width=1\linewidth]{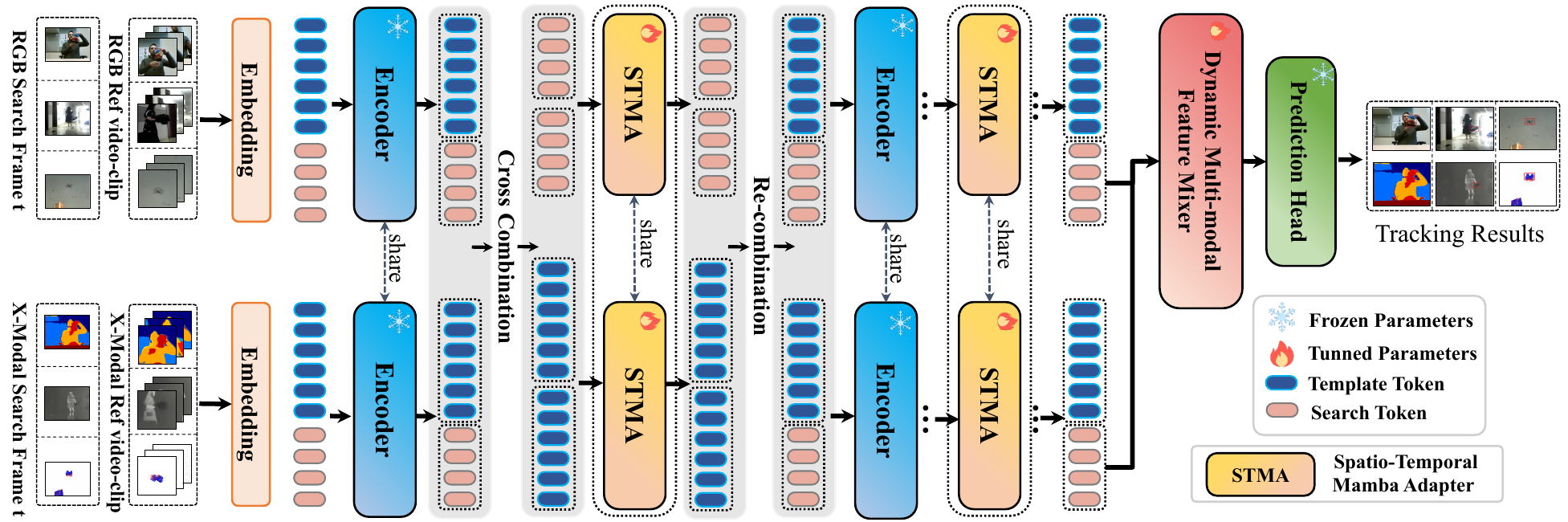}
    \end{center}
    \caption{\textbf{Overview architecture of {\tracker}.}
    {\tracker} freezes the encoder and performs adapter-based fine-tuning only on the proposed Spatio-Temporal Mamba Adapter (STMA) and Dynamic Multi-modal Feature Mixer (DMFM). It takes multimodal video clips as input for feature extraction, where X denotes infrared, depth, or event modalities. The learned multi-modal token features are then divided into two groups and fed into STMA to enable cross-modal interaction and spatio-temporal modeling. Finally, the RGB and X modality features are processed by DMFM to enhance their representational capacity, thereby achieving multi-modal object localization.
    }
    \label{fig:pic-02}
    \end{figure*}

\section{Approach}

This paper presents a novel multi-modal tracking method, {\tracker}, illustrated in Fig.\ref{fig:pic-02}. We first introduce the design principles of the state space model. Specifically, we design a Spatio-Temporal Mamba Adapter (STMA) that leverages Mamba’s long-sequence modeling capability to unify cross-modal interaction and spatio-temporal context learning. Subsequently, we propose a Dynamic Multi-modal Feature Mixer (DMFM) to further enhance the representational capacity of multi-modal features, thereby improving the discriminative power of the model.

\subsection{Preliminaries}
\label{sec:Preliminaries}

Gu et al. \cite{ssm2} propose the Structured State Space (S4) model, addressing the high computation and memory demands of the SSM. Since then, the SSM has become a promising architecture in deep sequence modeling. The models based on the SSM originate from a continuous system. This system maps the one-dimensional sequence \(x(t)\in\mathbb{R}^{L}\) to \(y(t)\in\mathbb{R}^{L}\) through the hidden state \(h(t)\in\mathbb{R}^{N}\). It defines the sequence-to-sequence transformation through the evolution parameter \(\boldsymbol{A}\in\mathbb{R}^{N\times N}\) and the projection parameters \(\boldsymbol{B}\in\mathbb{R}^{N\times1}\) and \(\boldsymbol{C}\in\mathbb{R}^{1\times N}\) as follows,
 \begin{equation}
        \begin{aligned}
            \left\{
            \begin{aligned}
                &h^{\prime}(t)=\boldsymbol{A}h(t)+\boldsymbol{B}x(t), \\
                &y(t)=\boldsymbol{C}h(t).
            \end{aligned}
            \right.
        \end{aligned}
    \end{equation}

    Modern SSMs (e.g., S4 and Mamba) are discrete forms of continuous state, discretized by zero-order hold (ZOH).
    
     \begin{equation}
        \begin{aligned}
            \left\{
            \begin{aligned}
                &\overline{\boldsymbol{A}}=\exp(\Delta{\boldsymbol{A}}), \\
                &\overline{\boldsymbol{B}}=(\Delta{\boldsymbol{A}})^{-1}(\exp(\Delta{\boldsymbol{A}}) - I)\cdot\Delta{\boldsymbol{B}}.  \\
                &h_t=\overline{\boldsymbol{A}}h_{t-1}+\overline{\boldsymbol{B}}x_t,  \\
                &y_t={\boldsymbol{C}}h_t.
            \end{aligned}
            \right.
        \end{aligned}
    \end{equation}
    The timescale parameter $\Delta$ is introduced, with \(h_t\) and \(h_{t - 1}\) representing discrete hidden states at different time steps. $\overline{\boldsymbol{A}}$ and $\overline{\boldsymbol{B}}$ are the discrete versions of parameters $\boldsymbol{A}$ and $\boldsymbol{B}$. 
    Mamba differs from previous time-invariant SSMs by incorporating the selective scan mechanism (S6), providing time-varying and context-aware features. The SSM parameters ${\boldsymbol{B}}\in\mathbb{R}^{B\times L\times N}$, ${\boldsymbol{C}}\in\mathbb{R}^{B\times L\times N}$, and $\Delta\in\mathbb{R}^{B\times L\times D}$ are parameterized using linear projection on input $x\in\mathbb{R}^{B\times L\times D}$.

\subsection{Overview}
\label{sec:overview}

As shown in Fig\ref{fig:pic-02}, our {\tracker} adopts a two-stream encoder structure with shared parameters for RGB-D, RGB-T, and RGB-E tracking. It consists of a Spatio-Temporal Mamba Adapter (STMA), a Dynamic Multi-modal Feature Mixer, and a prediction head. Specifically, {\tracker} encodes the search and template frames of RGB, depth, infrared, and event modalities into image patches respectively. 
The search and template tokens of each modality are concatenated to form the RGB token \(x_{0}^{V}\) and the token of other modalities \(x_{0}^{X}\). These tokens are input into the ViT module for feature extraction. As STMA (denoted as \(\mathcal{F}_{i}^{\text{STMA}}\)) is embedded in some intermediate layers for modal interaction. If on the \(i\)-th layer, it applies a cross-combination of the previous ViT layer's output to regroup tokens into \(x_{i}^{z}\) and \(x_{i}^{s}\). For them, STMA then performs unified modeling of cross-modal feature interaction and spatio-temporal cues, as described below.
     \begin{equation}
        \begin{aligned}
            \left\{
            \begin{aligned}
                &\hat{x}_{i}^{z}=\mathcal{F}_{i}^{\text{STMA}}(x_{i}^{z}),  \\
                &\hat{x}_{i}^{s}=\mathcal{F}_{i}^{\text{STMA}}(x_{i}^{s}), &i\in{2k\mid k\in\mathbb{N},1\leqslant k\leqslant 6},
            \end{aligned}
            \right.
        \end{aligned}
    \end{equation}
    where \(\hat{x}_{i}^{z}\) and \(\hat{x}_{i}^{s}\) represent the template and search features modeled by STMA, respectively. They are then recombined by modality into \(\hat{x}_{i}^{V}\) and \(\hat{x}_{i}^{X}\) for the next ViT layer. After all ViT layers and STMA blocks are done, we concatenate the search tokens of each modality along the channel dimension and input them into the  Dynamic Multi-modal Feature Mixer module (denoted as \(\mathcal{F}^{\text{DMFM}}\)) to get the final fusion feature token. The prediction head \(H\) then predicts tracking result \(B\) based on the output of the proposed mixer.

    \begin{equation}
        \begin{aligned}
                B=H(\mathcal{F}^{\text{DMFM}}(\hat{x}_{\text{N}}^{s})), & \quad \text{N}=12.
        \end{aligned}
    \end{equation}

\begin{figure}[t]
  \begin{center}
  \includegraphics[width=1\linewidth]{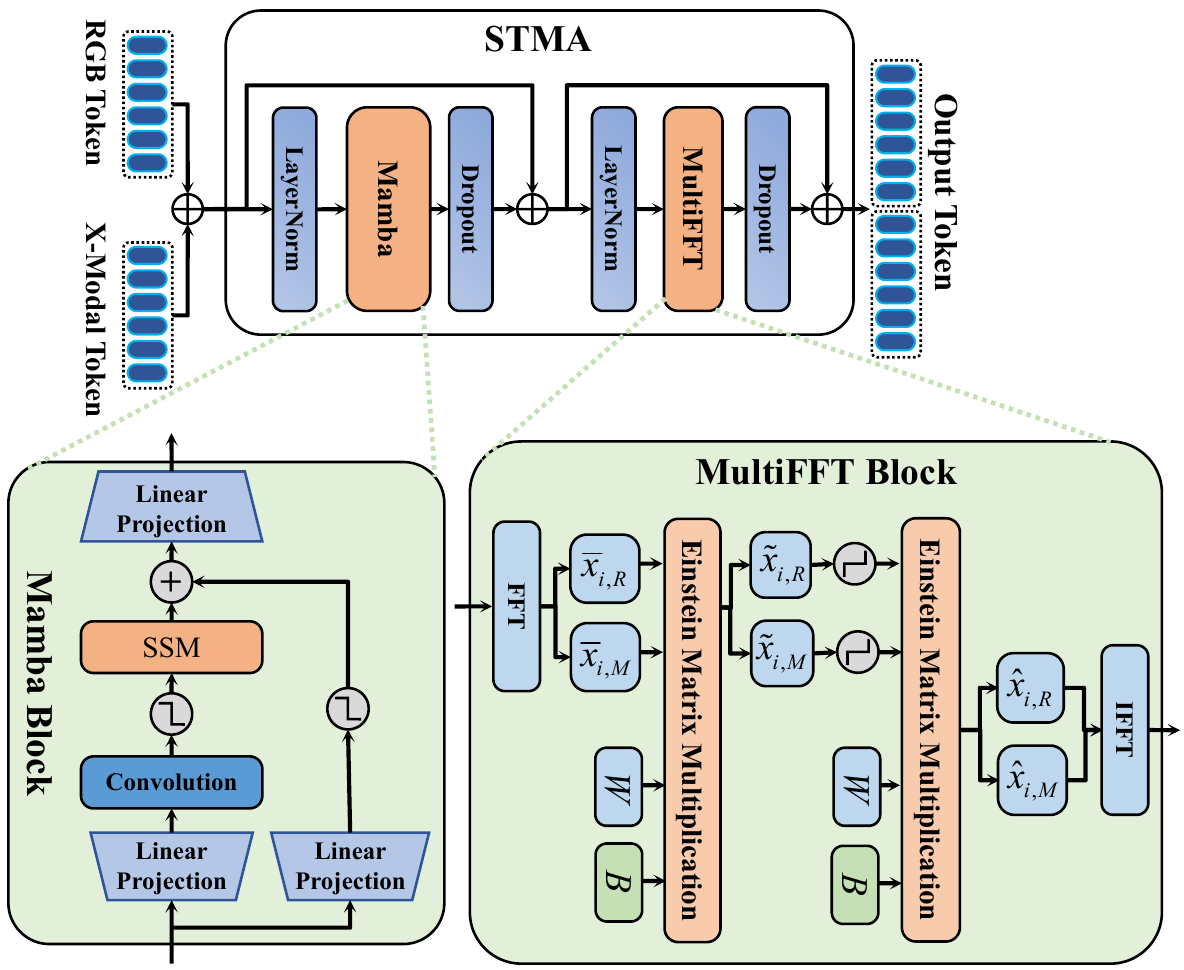}
  \end{center}
  \caption{
  Spatio-Temporal Mamba Adapter (STMA) combines cross-modal interactions and spatio-temporal cues for multi-modal tracking. It uses Mamba blocks for token modeling and MultiFFT for channel mixing. MultiFFT, composed of EMM, FFT and IFFT, to boost multi-modal feature extraction.
  }
  \label{fig:STMA}
\end{figure}

\subsection{Spatio-Temporal Mamba Adapter}
\label{sec:Spatio-Temporal Mamba Adapter}

We propose a simple and effective Spatio-Temporal Mamba Adapter (STMA) module for unified multi-modal tracking. 
The spatio-temporal mamba adapter plays a central role in unifying diverse modalities into a common representation, improving the efficiency and robustness of multi-modal fusion. It enables the combined modeling of cross-modal feature interactions and spatio-temporal cues. As shown in Fig.\ref{fig:STMA}, STMA includes Mamba blocks, MultiFFT, normalization, and dropout to enhance the tracker's ability to learn discriminative cross-modal features. it uses Mamba for multi-modal token modeling and MultiFFT for channel modeling. The simplified architectural details are as follows:
     \begin{equation}
        \begin{aligned}
            \left\{
            \begin{aligned}
                &\overline{x}_i=\text{DP}(\mathcal{F}^{\text{MBA}}(\text{Norm}(x_i))) + x_{i},  \\
                &\hat{x}_i=\text{DP}(\mathcal{F}^{\text{MFFT}}(\text{Norm}(\overline{x}_i))) + \overline{x}_i.
            \end{aligned}
            \right.
        \end{aligned}
    \end{equation}
    Here, \(\mathcal{F}^{\text{MBA}}\) is Mamba block, $i$ is the number of layers, {DP} refers to dropout, and {Norm} represents the normalization layer. The process normalizes the input \( x_i \), applies the Mamba block, and performs a residual connection and dropout. Then, the output feature \( \overline{x}_i \) is normalized and processed with MultiFFT (denoted as \(\mathcal{F}^{\text{MFFT}}\)) for frequency domain mixing. Finally, another dropout occurs, and the feature is added to the previous state. This structure applies to the \( i \)-th layer, with details as follows.
      
      The Mamba block models the input multi-modal sequence with linear complexity, improving the model's ability to learn complex cross-modal features and spatio-temporal context. The normalization layer ({Norm}) enhances stability and convergence during training.
     
     MultiFFT uses Einstein matrix multiplication (EMM), an efficient feature channel mixing operation that boosts computation and feature extraction. It mixes channels of tokens modeled by Mamba, enhancing global visibility and energy concentration in complex multi-modal data. Below are the details of how MultiFFT processes \(\overline{x}_i\) at \(i\)-th layer.

\begin{figure}[t]
  \begin{center}
  \includegraphics[width=1\linewidth]{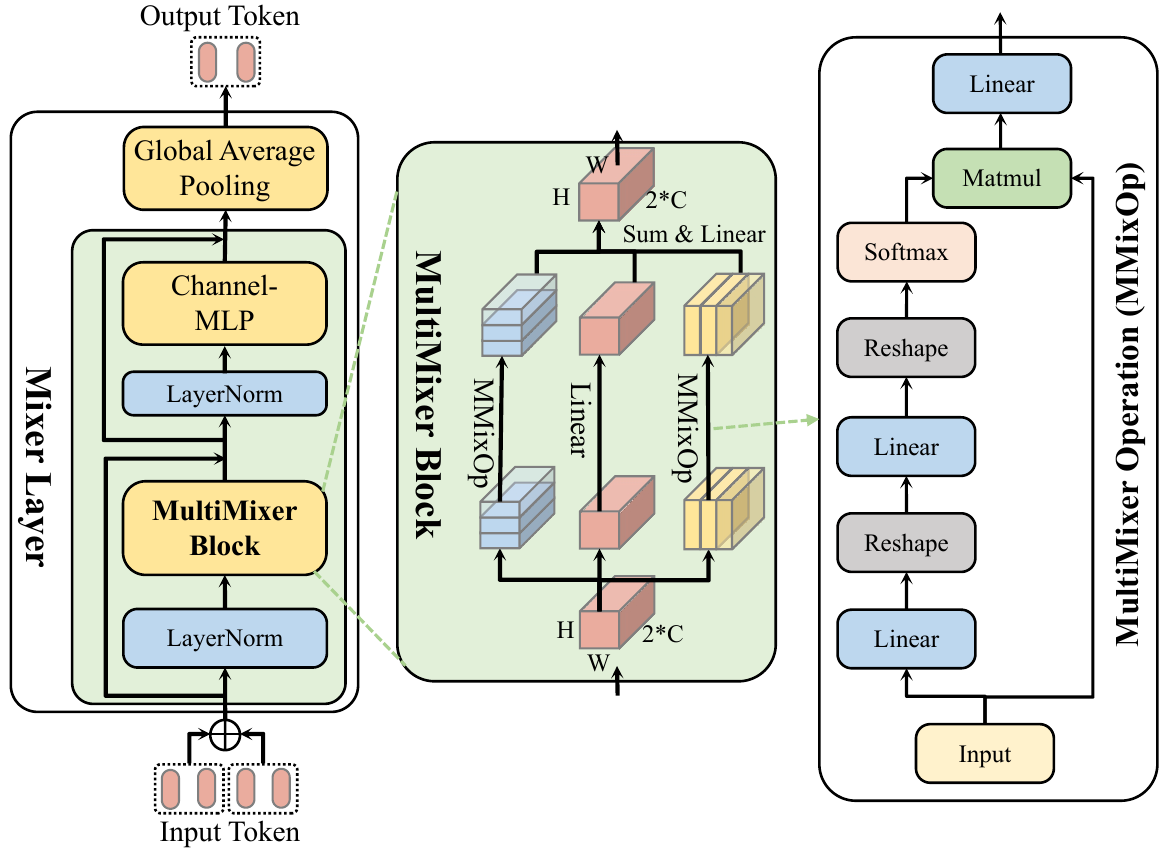}
  \end{center}
  \caption{
  Dynamic Multi-modal Feature Mixer fuses cross-modal features using a Mixer Layer, Global Average Pooling, and Channel-MLP. The Mixer Layer combines token and channel information, while MMixOp enhances feature fusion.
  }
  \label{fig:DMFM}
\end{figure}

     \begin{equation}
        \begin{aligned}
            \left\{
            \begin{aligned}
                &(\overline{x}_{i,R}  , \overline{x}_{i,M}) = \text{FFT}(\overline{x}_{i}),\\
                &(\widetilde{x}_{i,R}  , \widetilde{x}_{i,M}) = \text{EMM}\left((\overline{x}_{i,R}  , \overline{x}_{i,M}), \mathcal{W}, \mathcal{B}\right), \\
                &{y}_{i} = ({y}_{i,R}, {y}_{i,M}) = \sigma(\widetilde{x}_{i,R} ,\widetilde{x}_{i,M}), \\
                &(\hat{x}_{i,R} ,\hat{x}_{i,M}) = \text{EMM}\left({y}_{i}, \mathcal{W}, \mathcal{B}\right), \\
                &\hat{x}_{i} = \text{IFFT}(\hat{x}_{i,R} ,\hat{x}_{i,M}).
            \end{aligned}
            \right.
        \end{aligned}
    \end{equation}
    
    Here, the Fast Fourier Transform ({FFT}) and Inverse Fast Fourier Transform ({IFFT}) work along the channel dimension. The {FFT} transforms the input from the physical domain to the frequency domain, while {IFFT} does the opposite. MultiFFT first applies {FFT} to \(\overline{x}_{i}\), producing \((\overline{x}_{i,R}, \overline{x}_{i,M})\). It then uses Einstein Matrix Multiplication ({EMM}) to reduce complexity and optimizes tokens with the sigmoid function \(\sigma\). Finally, the output \(\hat{x}_i\) is obtained via {IFFT}. In {EMM}, \(\mathcal{W}\in\mathbb{R}^{C_b \times C_d \times C_d}\) is the complex weight matrix, \(\mathcal{B}\) is the bias, and \(\mathcal{W} = \mathcal{W}_R + \mathcal{W}_M\).
    
    Through the proposed spatio-temporal mamba adapter, we inject multi-modal information (including infrared, depth, and event modalities) and temporal contextual cues into the model, thereby enhancing its cross-modal complementary perception and improving its sensitivity to temporal cues. This design facilitates more accurate target localization and tracking across diverse multi-modal scenarios (see Fig.\ref{fig:drawtrackers} and Fig.\ref{fig:drawattn}).

\subsection{Dynamic Multi-modal Feature Mixer}

The Dynamic Multi-modal Feature Mixer (DMFM) consists of a Mixer Layer, Global Average Pooling ({GAP}), and Channel-MLP. It performs multi-stage fusion and dimensionality reduction to enhance the discriminative power of cross-modal target information. As shown in Fig.\ref{fig:DMFM}, DMFM takes the input $\hat{x}_{\text{N}}^{s}$ from the encoders, passes it through the Mixer Layer to generate output tokens \(b\), and then applies Global Average Pooling. The details of DMFM are as follows.

     \begin{equation}
        \begin{aligned}
            \begin{aligned}
                & \hat{b} = \mathcal{F}^{\text{DMFM}}(\hat{x}_{\text{N}}^{s}) =\text{GAP}(\mathcal{F}^{\text{Mix}}(\hat{x}_{\text{N}}^{s})).  \\
            \end{aligned}
        \end{aligned}
    \end{equation}

    The Mixer Layer (denoted as $\mathcal{F}^{\text{Mix}}$) effectively combines token and channel information to enhance the model's ability to learn cross-modal features. It consists of two key components: the MultiMixer Block ($\mathcal{F}^{\text{MixB}}$) and the Channel-MLP Block ($\mathcal{F}^{\text{CMLP}}$). The details of $\mathcal{F}^{\text{Mix}}$ are as follows.

     \begin{equation}
        \begin{aligned}
            \left\{
            \begin{aligned}
                & s=\mathcal{F}^{\text{MixB}}(\text{Norm}(\hat{x}_{\text{N}}^{s})) + \hat{x}_{\text{N}}^{s},  \\
                & b=\mathcal{F}^{\text{CMLP}}(\text{Norm}(s)) + s.
            \end{aligned}
            \right.
        \end{aligned}
    \end{equation}

    The MultiMixer Block uses the MultiMixer Operation (MMixOp). It performs column and row mixing (using MMixOp to generate dynamic matrices for multi-modal token fusion) and channel mixing (through linear transformation). This results in \(s_w\), \(s_h\), and \(s_c\), which are summed and linearly transformed with \(\psi\) to produce the final output: 

    \begin{equation}
        \begin{aligned}
                s = \psi(s_w + s_h + s_c).
        \end{aligned}
    \end{equation}This process effectively fuses multi-modal token information, with MMixOp providing the core dynamic mixing capability.
    Specifically, MMixOp divides the features into segments, reduces the dimensionality of each segment, and computes its mixing matrix. It then mixes these results to obtain \(s_w\), \(s_h\), and \(s_c\), which are combined via \(\psi\) to form the final output. It reduces computational complexity and enhances the model's adaptability to multi-modal features.
    
    The Channel-MLP Block is a feed-forward layer, like in transformer, that fuses channel information to improve the model's handling of multi-modal data.

\subsection{Update, Prediction head and Loss function}

\textbf{Update}. Unlike methods that retrain classifiers for foreground quality \cite{Stark} or use complex thresholding strategies \cite{TATrack}, we apply a simple and effective template update strategy. Similar to ODTrack \cite{odtrack}, the template index is determined using the formula:
     \begin{equation}
        \begin{aligned}
            \left\{
            \begin{aligned}
                &\{0\} \cup \{i\cdot T+\left\lfloor\frac{T}{2}\right\rfloor\mid i\in\{0,1,\ldots,M - 1\}\}, &\text{if }M>1 \\
                &\{0\}, &\text{if }M = 1
            \end{aligned}
            \right.
        \end{aligned}
    \end{equation}
Where $T=\left\lfloor\frac{C_i}{M}\right\rfloor$ denotes the average memory duration of each template. ${C_i}$ is the current frame index. This formulation determines that, during the inference phase of the entire video sequence, online template frames are sampled at uniform time intervals, effectively preserving important target contextual cues from both the initial and online templates, and facilitating long-term cross-frame contextual integration.

    \textbf{Prediction head}. For prediction head network design, we utilize traditional classification and bounding box regression heads. The classification score map $\mathbb{R}^{1\times\frac{H_s}{p}\times\frac{W_s}{p}}$, bounding box size $\mathbb{R}^{2\times\frac{H_s}{p}\times\frac{W_s}{p}}$, and offset size $\mathbb{R}^{2\times\frac{H_s}{p}\times\frac{W_s}{p}}$ are derived from three sub-convolutional networks.

    \textbf{Loss function}. We use focal loss \cite{ross2017focal} as $L_{cls}$, $L_1$ loss and GIoU loss \cite{rezatofighi} for regression. Total loss $L$ is: 

    \begin{equation}
        \begin{aligned}
                L=L_{cls}+\lambda_1L_1+\lambda_2L_{GIoU}.
        \end{aligned}
    \end{equation} Using $\lambda_1 = 5$ and $\lambda_2 = 2$ as regularization parameters. For video segment modeling, each frame’s task loss is independent, with the final loss averaged over all search frames. 


\begin{table}[t]
    \centering

    \caption{Performance comparisons with state-of-the-art trackers on RGB-T tracking. The top three results are shown in {\protect\color{red}red}, {\protect\color{blue}blue}, and {\protect\color{green}green}, respectively.}
    \label{tab:RGB+Thermal}
    \resizebox{\linewidth}{!}{
    \begin{tabular}{c|ccc|cc|cc}
    \toprule
    \multicolumn{1}{c|}{\multirow{2}{*}{Method}}
    & \multicolumn{3}{c|}{LasHeR \cite{LasHeR}} 
    & \multicolumn{2}{c|}{RGBT234 \cite{RGBT234}}  
    & \multicolumn{2}{c}{RGBT210 \cite{RGBT210}} \\ 
    \cline{2-8}

    & SR & PR & NPR & MSR & MPR & MSR & MPR\\ 
    \midrule
    {\tracker}-384 &\textbf{\color{red}{60.1}} & \textbf{\color{red}{76.0}} & \textbf{\color{red}{72.2}} & \textbf{\color{red}{70.1}} & \textbf{\color{red}{92.6}} & \textbf{\color{red}{67.9}} & \textbf{\color{red}{91.2}}  \\
    {\tracker}-256 &\textbf{\color{blue}{58.2}} & \textbf{\color{blue}{73.5}} & \textbf{\color{blue}{69.9}} & \textbf{\color{blue}{68.6}} & \textbf{\color{blue}{90.8}} & \textbf{\color{blue}{66.2}} & \textbf{\color{blue}{89.6}}  \\
    \midrule
    OneTracker \cite{OneTracker} &53.8 &67.2 & - &64.2 &85.7 & - & -  \\
    SDSTrack \cite{SDSTrack} &53.1 &66.5 & - &62.5 &84.8 & - & -  \\
    Un-Track \cite{un-track} & 51.3  & 63.7 & 60.0 &62.5 &84.2 & - & -  \\
    TATrack \cite{TATrack} &56.1 &70.2 &66.7 &64.4 &87.2 &61.8 & \textbf{\color{green}85.3}  \\
    BAT \cite{BAT} &56.3 &70.2 &66.4 &64.1 &86.8 & - & -  \\
    GMMT \cite{gmmt} & \textbf{\color{green}56.6} & \textbf{\color{green}70.7} & \textbf{\color{green}67.0} & \textbf{\color{green}64.7} &\textbf{\color{green}{87.9}} & - & -  \\
    TBSI \cite{TBSI} &56.3 &70.5 &66.6 &64.3 &86.4 & \textbf{\color{green}62.5} & \textbf{\color{green}85.3}  \\
    ViPT \cite{ViPT} &52.5 &65.1 &61.7 &61.7 &83.5 & 59.2 & 81.3  \\

    OSTrack \cite{OSTrack} &41.2 &51.5 & - &54.9 &72.9 & - & -  \\
    
    APFNet \cite{APFNet} & 36.2 & 50.0 & 43.9 & 57.9 & 82.7 & - & -  \\
    ProTrack \cite{protrack} &53.8 &42.0 &49.8 &59.9 &79.5 & - & -  \\
    DMCNet \cite{DMCNet} & 35.5 & 49.0 & 43.1  & 59.3 & 83.9 & 55.5 & 79.7   \\
    FANet \cite{FANet} & 30.9 & 44.1 & 38.4 & 55.3  & 78.7  & - & -  \\
    CAT \cite{CAT}  & 31.4 & 45.0 & 39.5 & 56.1  & 80.4 & 53.3 & 79.2   \\
    mfDiMP \cite{mfDiMP} & 34.3 & 44.7 & 39.5 & 42.8 &64.6  & 55.5 & 78.6   \\
    \bottomrule    
    \end{tabular} }
\end{table}

\begin{table}[t]
    
    \centering
    \caption{Performance comparisons with state-of-the-art trackers on RGB-D tracking. The top three results are shown in {\protect\color{red}red}, {\protect\color{blue}blue}, and {\protect\color{green}green}, respectively.}
    \label{tab:RGB+DEPTH}
    \resizebox{\linewidth}{!}{
    \begin{tabular}{c|ccc|ccc}
    \toprule
    \multicolumn{1}{c|}{\multirow{2}{*}{Method}}
    & \multicolumn{3}{c|}{DepthTrack \cite{DeT}} 
    & \multicolumn{3}{c}{VOT-RGBD22 \cite{DMTrack}} \\ 
    \cline{2-7}

    & F-score & Re & Pr & EAO & Acc. & Rob.\\ 
    \midrule
    {\tracker}-384 &\textbf{\color{red}{67.3}} & \textbf{\color{red}{66.9}} & \textbf{\color{red}{67.7}} & \textbf{\color{red}{77.8}} & \textbf{\color{red}{82.5}} & \textbf{\color{red}{93.7}}  \\
    {\tracker}-256 &\textbf{\color{blue}{63.5}} & \textbf{\color{blue}{63.7}} & \textbf{\color{blue}{63.3}} & \textbf{\color{blue}{77.2}} &81.4 & \textbf{\color{blue}{93.2}} \\
    \midrule
    OneTracker \cite{OneTracker} &60.9 &60.4 & 60.7 &72.7 &\textbf{\color{green}{81.9}} & 87.2  \\
    SDSTrack \cite{SDSTrack} &\textbf{\color{green}{61.4}} &\textbf{\color{green}{60.9}} &\textbf{\color{green}{61.9}} &\textbf{\color{green}{72.8}} &81.2 &\textbf{\color{green}{88.3}}  \\
    Un-Track \cite{un-track} &61.0 &60.8 &61.1 &72.1 &\textbf{\color{blue}{82.0}} &86.9  \\
    ViPT \cite{ViPT} &59.4 &59.6 &59.2 &72.1 &81.5 &87.1  \\
    ProTrack \cite{protrack} &57.8 &57.3 &58.3 &65.1 &80.1 &80.2  \\
    SPT \cite{SPT} &53.8 &54.9 &52.7 &65.1 &79.8 &85.1  \\
    SBT-RGBD \cite{SBT} &- &- &- &70.8 &80.9 &86.4  \\
    OSTrack \cite{OSTrack} &52.9 &52.2 &53.6 &67.6 &80.3 &83.3  \\
    DeT \cite{DeT} &53.2 &50.6 &56.0 &65.7 &76.0 &84.5  \\
    DMTrack \cite{DMTrack} &- &- &- &65.8 &75.8 &85.1  \\
    DDiMP \cite{DDiMP} &48.5 &56.9 &50.3 &- &- &-  \\
    STARK-RGBD \cite{Stark} &- &- &- &64.7 &80.3 &79.8  \\
    KeepTrack \cite{KeepTrack} &- &- &- &60.6 &75.3 &79.7  \\
    ATCAIS \cite{DDiMP} &47.6 &45.5 &50.0 &55.9 &76.1 &73.9  \\
    LTMU-B \cite{LTMU} &46.0 &41.7 &51.2 &- &- &-  \\
    Siam-LTD \cite{DDiMP} &37.6 &34.2 &41.8 &- &- &-  \\
    \bottomrule    
    \end{tabular}
    }
\end{table}

 \begin{table*}[t]
    \centering
    \caption{Performance comparisons with state-of-the-art trackers on RGB-E tracking. The top three results are shown in {\protect\color{red}red}, {\protect\color{blue}blue}, and {\protect\color{green}green}, respectively.}
    \label{tab:rgb_e_tracking}
    \resizebox{\textwidth}{!}{ 
    \begin{tabular}{c|ccccccccccc|cc}
        \midrule
        & STARK\_E & PrDiMP\_E & LTMU\_E & TransT\_E & SiamRCNN\_E & OSTrack & ProTrack & ViPT & Un-Track & SDSTrack & OneTracker 
        & {\tracker} & {\tracker} \\
        & \makecell{\protect\cite{Stark}} &\makecell{\protect\cite{PrDiMP}} &\makecell{\protect\cite{LTMU}} &\makecell{\protect\cite{TransT}} &\makecell{\protect\cite{SiamRCNN}} &\makecell{\protect\cite{OSTrack}} &\makecell{\protect\cite{protrack}} &\makecell{\protect\cite{ViPT}} & \makecell{\protect\cite{un-track}} &\makecell{\protect\cite{SDSTrack}} &\makecell{\protect\cite{OneTracker}} 
        & \makecell{-256} &\makecell{-384}\\
        \midrule
        MPR & 61.2 & 64.4 & 65.5 & 65.0 & 65.9 & 69.5 & 63.2 & 75.8 & 75.5 & {\color{green}76.7} & {\color{green}76.7} & {\color{blue}77.3} & {\color{red}\textbf{79.7}} \\
        MSR & 44.6 & 45.3 & 45.9 & 47.4 & 49.9 & 53.4 & 47.1 & 59.2 & 58.9 & 59.7 & {\color{blue}60.8} & {\color{green}60.4} & {\color{red}\textbf{62.7}} \\ 
        \midrule
    \end{tabular}
    }
\end{table*}

\begin{table}[t]
    \centering
     \caption{Comparison of performance, model parameters, and inference speed on LasHeR.}
    \label{tab:speed_compare}
     \resizebox{\linewidth}{!}{
    \begin{tabular}{c|cccc}
    \toprule
    Method & Success & Learnable Parameters (M) & Speed (FPS) & Device\\ 
    \midrule
    \textrm{{\tracker}-256} & 58.2 & 11.9 & 32.5 & V100 \\
    \textrm{{\tracker}-384} & 60.1 & 11.9 & 18.4 & V100 \\
    \textrm{SDStrack \cite{SDSTrack}} & 53.1 & 14.8 & 17.9 & V100 \\
    \textrm{UnTrack \cite{un-track}} & 51.3 & 6.2 & 19.5 & V100 \\
    \bottomrule    
    \end{tabular}
    }
\end{table}

\section{Experiments}
\label{sec:Experiments}

\subsection{Implementation Details}
    \textbf{Model Variants}. Two variants of {\tracker} with diverse configurations are trained by us as follows. {\tracker}-256 has the template size [128×128] and search region size [256×256]; {\tracker}-384 has the template size [192×192] and search region size [384×384].

    \textbf{Training}. We propose {\tracker}, which handles multiple modality combinations in a single training phase. Using OSTrack \cite{OSTrack} as the foundation tracker, {\tracker}-256 and {\tracker}-384 are trained on PyTorch with batch sizes of 16 and 8, respectively, using 4 NVIDIA RTX A800 GPUs. Input data consists of three reference frames and two search frames. During training, we freeze all foundation tracker parameters, fine-tuning only the Spatio-Temporal Mamba Adapter and the Dynamic Multi-modal Feature Mixer. We sample from LasHeR \cite{LasHeR}, DepthTrack \cite{DeT}, and VisEvent \cite{VisEvent} datasets at a 1:1:1 ratio for 15 epochs, with each epoch containing 60k image pairs. The training follows the foundation tracker's loss function parameters, using the AdamW optimizer with weight decay of 1e-4 and an initial learning rate of $2 \times 1e-4$, which decays by a factor of 10 after 10 epochs.

    \textbf{Inference}. For consistency with training settings, we use three equally spaced reference frames in inference. Search frames are input frame by frame.

\begin{figure}[t]
  \begin{center}
  \includegraphics[width=1\linewidth]{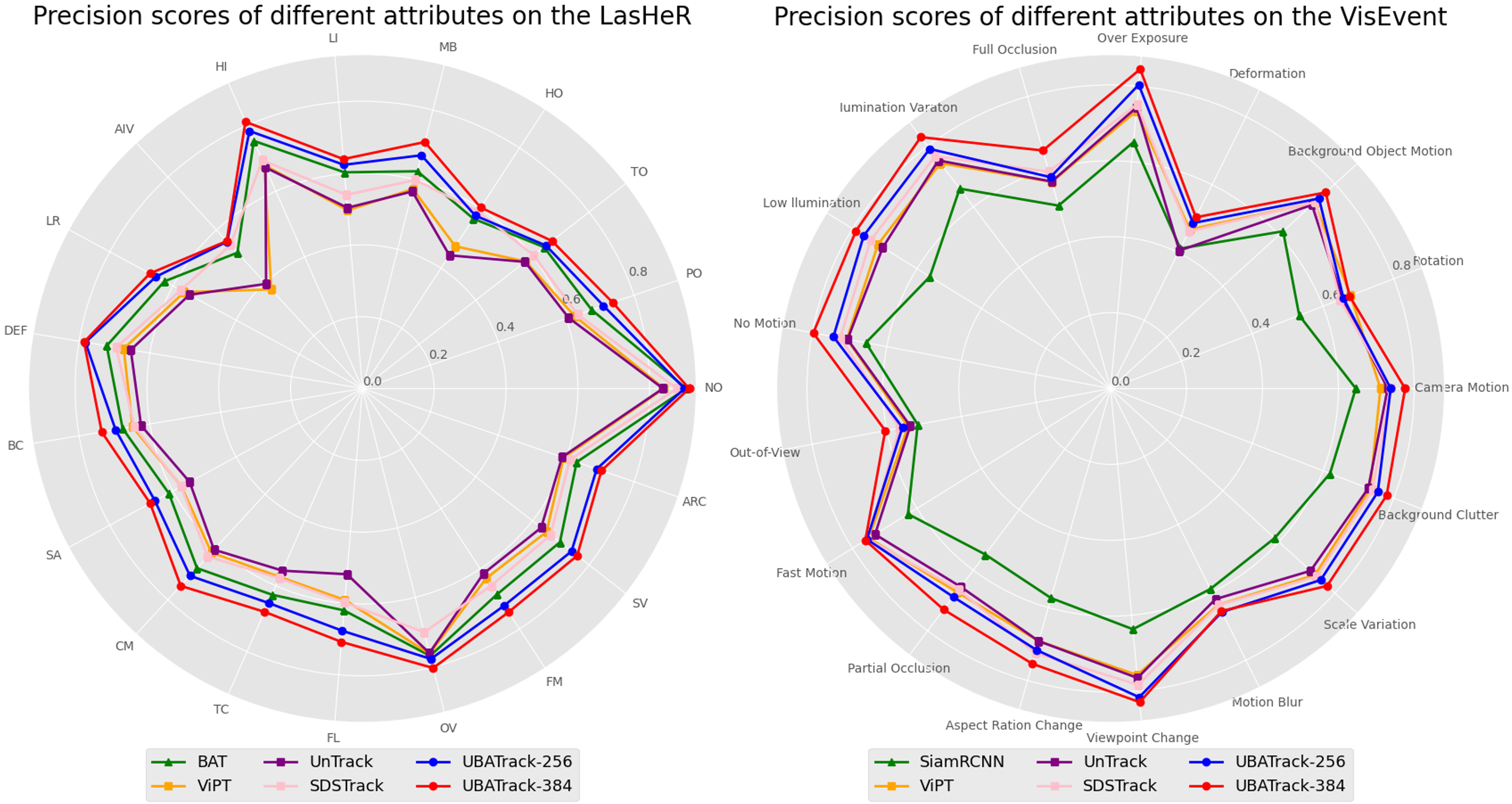}
  \end{center}
  \caption{Precision scores of different attributes on the LasHeR \cite{LasHeR} and VisEvent \cite{VisEvent} dataset.}
  \label{fig:attributes}
\end{figure}

\subsection{Comparison with State-of-the-Arts}

     \textbf{LasHeR}. LasHeR \cite{LasHeR} is currently the largest RGB-T tracking dataset with precise annotation and alignment. As illustrated in Tab.\ref{tab:RGB+Thermal}, {\tracker}-256 outperforms all previous SOTA trackers. It attains the leading performance with precision and success rates of 73.5\% and 58.2\% respectively. Notably, the {\tracker}-384's precision exceeds the previous optimal result by 4.4\%.

     \textbf{RGBT234}. The RGBT234 \cite{RGBT234} dataset comprises 234 sequences and approximately 116.7K frames. 
     MSR and MPR are utilized for performance evaluation. As shown in Tab.\ref{tab:RGB+Thermal}, {\tracker}-256 achieves 90.8\% MPR and 68.6\% MSR, surpassing the MPR and MSR scores of the existing SOTA GMMT \cite{gmmt} by 3.9\% and 2.9\%, respectively. Additionally, {\tracker}-384 outperforms {\tracker}-256 by 1.8\% and 1.5\% on the same metrics.

    \textbf{RGBT210}. The RGBT210 \cite{RGBT210} dataset comprises 210 sequences with approximately 104.8K frames.
    In Tab.\ref{tab:RGB+Thermal},
    {\tracker}-384 achieves the best performance in terms of MPR and MSR metrics on this dataset, which are 91.2\% and 67.9\% respectively. The {\tracker}-256 also surpasses the MPR and MSR scores of the existing SOTA TBSI \cite{TBSI} by 3.7\% and 4.3\%, respectively.

\begin{table}[t]
    \centering
    \caption{Ablation studies of each component in {\tracker}.}
    \label{tab:ablationtabs4}
    \resizebox{\linewidth}{!}{
    \begin{tabular}{c|cc|ccc|cc|ccc}
    \toprule
    \multicolumn{1}{c|}{\multirow{2}{*}{Variants}}
    & \multicolumn{2}{c|}{Method} 
    & \multicolumn{3}{c|}{LasHeR \cite{LasHeR}} 
    & \multicolumn{2}{c|}{VisEvent \cite{VisEvent}}  
    & \multicolumn{3}{c}{DepthTrack \cite{DeT}} \\ 
    \cline{2-11}
    
    & DMFM & STMA & SR & PR & NPR & MSR & MPR & F-score & Re & Pr\\ 
    \midrule
    Baseline & & &52.5 &66.0 &62.1 &57.6 &75.1 &55.9 &55.9 &55.9 \\
    \midrule
    \textrm{I}  & $\checkmark$ & &54.3 &68.9 &65.0 &58.3 &75.7 &59.9 &59.4 &60.5 \\
    \textrm{II}  & & $\checkmark$ &57.5 &72.3 &69.0 &60.4 &76.9 &61.5 &61.8 &61.2 \\
    \midrule
    \textrm{III}  &$\checkmark$ &$\checkmark$ &\textbf{58.2} &\textbf{73.5} &\textbf{69.9} &\textbf{60.4} &\textbf{77.3} &\textbf{63.5} &\textbf{63.7} &\textbf{63.3} \\
    \bottomrule    
    \end{tabular}
    } 
\end{table}

\begin{table}[t]
    \centering
    \caption{Ablation studies on the Spatio-Temporal Mamba Adapter and other modal interaction methods.}
    \label{tab:ablationtabs5}
    \resizebox{\linewidth}{!}{
    \begin{tabular}{c|c|ccc|cc|ccc}
    \toprule
    \multicolumn{1}{c|}{\multirow{2}{*}{Variants}}
    & \multicolumn{1}{c|}{\multirow{2}{*}{\# Params}}
    & \multicolumn{3}{c|}{LasHeR \cite{LasHeR}} 
    & \multicolumn{2}{c|}
    {VisEvent \cite{VisEvent}}
    & \multicolumn{3}{c}{DepthTrack \cite{DeT}} \\ 
    \cline{3-10}
    
    & & SR & PR & NPR & MSR & MPR & F-score & Re & Pr\\ 
    \midrule
    Baseline &- &52.5 &66.0 &62.1 &57.6 &75.1 &55.9 &55.9 &55.9 \\
    \midrule  
    Attention-based &42.527M &57.1 &71.3 &68.3 &59.1 &76.2 &61.1 &61.5 &60.8 \\
    \midrule
    Mamba-based &0.016M &56.2 &71.2 &67.6 &59.8 &76.6 &58.4 &58.3 &58.6 \\
    Mamba with MLP &42.794M &56.2 &71.0 &67.5 &58.1 &75.6 &61.1 &61.6 &60.6 \\
    Mamba with MultiFFT &0.018M &\textbf{57.5} &\textbf{72.3} &\textbf{69.0} &\textbf{60.4} &\textbf{76.9} &\textbf{61.5} &\textbf{61.8} &\textbf{61.2} \\
    \bottomrule 
    \end{tabular}
    }
\end{table}

    \textbf{DepthTrack}. DepthTrack \cite{DeT} serves as a comprehensive RGBD tracking benchmark, featuring 150 training and 50 testing sequences and evaluation metrics such as precision (Pr), recall (Re), and F-score. As shown in Tab.\ref{tab:RGB+DEPTH}, {\tracker}-256 surpasses the precision, recall and F-score of the existing SOTA tracker's SDSTrack \cite{SDSTrack} by 1.4\%, 2.8\% and 2.1\%, respectively. {\tracker}-384 achieves new SOTA performance, in terms of Pr, Re and F-score on this dataset, which are 67.7\%, 66.9\% and 67.3\% respectively.

    \textbf{VOT-RGBD2022}. VOT-RGBD2022 \cite{DMTrack} is the recent dataset in RGB-D tracking, consisting of 127 short-term RGB-D sequences. Performance is evaluated by Accuracy, Robustness, and Expected Average Overlap (EAO), with EAO as the main metric. In Tab.\ref{tab:RGB+DEPTH}, our {\tracker} models outperform all previous RGB+Depth tracking methods. {\tracker}-384 achieves the highest EAO score of 77.8\%.

    \textbf{VisEvent}. VisEvent \cite{VisEvent} represents the largest dataset dedicated to RGB-E tracking. It comprises 500 pairs of videos for training and 320 pairs for testing. In Tab.\ref{tab:rgb_e_tracking}, {\tracker}-256 surpasses the MPR scores of the existing SOTA OneTracker \cite{OneTracker} by 0.6\%. {\tracker}-384 establishes the best MPR score of 79.7\%, outperforming the OneTracker by 3\%.

\begin{figure}[t]
    \begin{center}
    \includegraphics[width=1\linewidth]{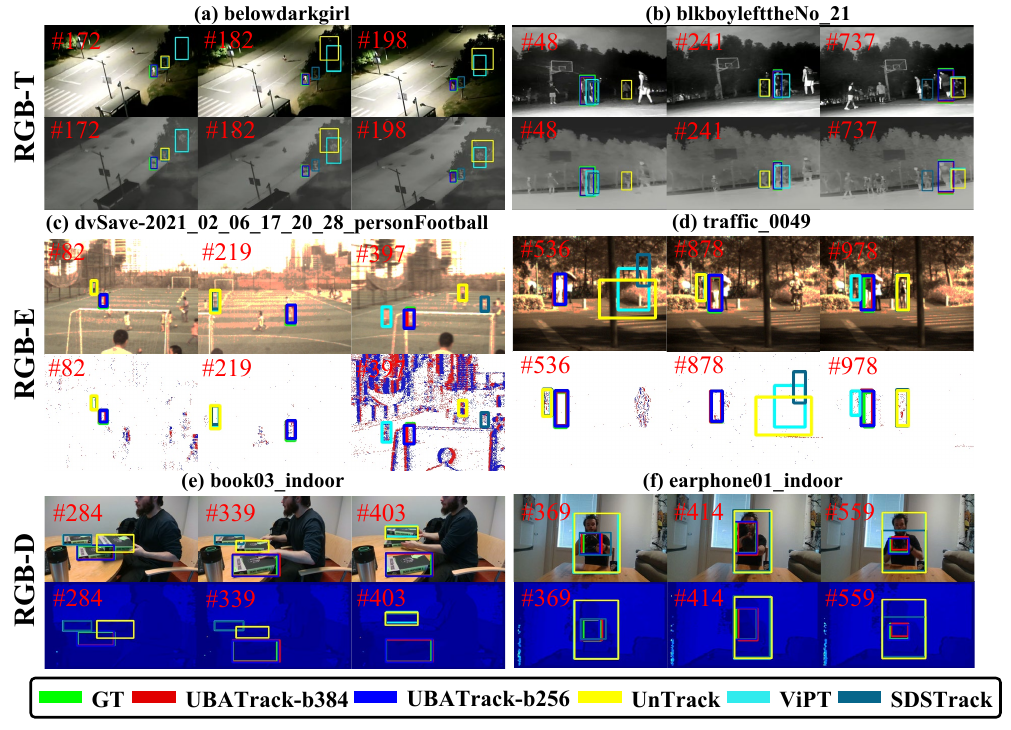}
    \end{center}
    \caption{Qualitative comparison of {\tracker} with other trackers using LasHeR \cite{LasHeR}, VisEvent \cite{VisEvent} and DepthTrack \cite{DeT} dataset.}
    \label{fig:drawtrackers}
\end{figure}

\begin{table}[t]
    \centering
    \caption{Ablation studies on the the number of STMA Blocks.}
    \label{tab:ablationtabs6}
    \resizebox{\linewidth}{!}{
    \begin{tabular}{c|c|ccc|cc|ccc|c}
    \toprule
    \multicolumn{1}{c|}{\multirow{2}{*}{Number}}
    & \multicolumn{1}{c|}{\multirow{2}{*}{\# Params}}
    & \multicolumn{3}{c|}{LasHeR \cite{LasHeR}} 
    & \multicolumn{2}{c|}{VisEvent \cite{VisEvent}}  
    & \multicolumn{3}{c|}{DepthTrack \cite{DeT}} & \multicolumn{1}{c}{\multirow{2}{*}{FPS}} \\ 
    \cline{3-10}
    & & SR & PR & NPR & MSR & MPR & F-score & Re & Pr\\ 
    \midrule
    0 &- &54.3 &68.9 &65.0 &58.3 &75.7 &59.9 &59.4 &60.5 &\textbf{40.3}\\
    1 &0.003M &56.7 &71.6 &68.0 &59.8 &76.8 &61.3 &61.4 &61.1 &37.1\\
    2 &0.006M &57.2 &72.4 &68.7 &60.2 &77.1 &61.1 &61.2 &61.0 &34.0\\
    4 &0.012M &57.8 &73.1 &69.5 &60.1 &77.0 &61.9 &61.8 &61.9 &29.6\\
    6 &0.018M &\textbf{58.2} &\textbf{73.5} &\textbf{69.9} &\textbf{60.4} &\textbf{77.3} &\textbf{63.5} &\textbf{63.7} &\textbf{63.3} &26.5\\
    12 &0.034M &57.5 &72.5 &69.1 &60.1 &77.0 &58.5 &58.2 &58.8 &18.4\\
    \bottomrule    
    \end{tabular}
    }
\end{table}

\subsection{Exploration Study}

    \textbf{Component Analysis}. To analyze the effectiveness of {\tracker}’s components, we create four ablation variants in Tab.\ref{tab:ablationtabs4}. The Baseline uses an RGB tracker \cite{OSTrack} as the foundation. The \textrm{I} adds the DMFM module to Baseline. The \textrm{II} adds the STMA module to Baseline. The \textrm{III} is {\tracker}-256 itself. Results in Tab.\ref{tab:ablationtabs4} (on the tri-modal dataset) show: (a) Comparing Baseline and the \textrm{I}, the \textrm{I} improves performance, indicating DMFM helps learn better fused search region features. (b) The \textrm{II} significantly outperforms Baseline, confirming STMA's effectiveness. (c) Combining DMFM and STMA in the \textrm{III} surpasses all other variants, demonstrating each component’s effectiveness and their combined impact on performance.

\begin{figure}[t]
    \begin{center}
    \includegraphics[width=1\linewidth]{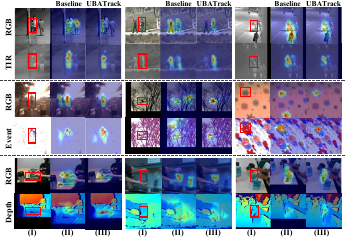}
    \end{center}
    \caption{Attention maps of the search regions in the RGB and other modalities for the baseline tracker (in Tab.\protect\ref{tab:ablationtabs4}) and {\tracker} are visualized from the RGB234 \cite{RGBT234}, VisEvent \cite{VisEvent} and DepthTrack \cite{DeT} dataset. The figure includes: (\textrm{I}) Original frames with highlighted tracked objects. (\textrm{II}) Attention maps in the baseline tracker. (\textrm{III}) Attention maps in {\tracker}.}
  \label{fig:drawattn}
\end{figure}

\begin{table}[t]
    \centering
    \caption{Ablation studies on the Dynamic Multi-modal Feature Mixer.}
    \label{tab:ablationtabs7}
    \resizebox{\linewidth}{!}{
    \begin{tabular}{c|ccc|cc|ccc}
    \toprule
    \multicolumn{1}{c|}{\multirow{2}{*}{Variants}}
    & \multicolumn{3}{c|}{LasHeR \cite{LasHeR}} 
    & \multicolumn{2}{c|}
    {VisEvent \cite{VisEvent}}
    & \multicolumn{3}{c}{DepthTrack \cite{DeT}} \\ 
    \cline{2-9}
    
    & SR & PR & NPR & MSR & MPR & F-score & Re & Pr\\ 
    \midrule
    Baseline &52.5 &66.0 &62.1 &57.6 &75.1 &55.9 &55.9 &55.9 \\
    \midrule
    add conv &53.0 &66.7 &63.2 &57.3 &75.0 &57.5 &57.5 &57.5 \\
    add MLP &52.4 &65.9 &62.2 &57.7 &75.1 &59.6 &59.3 &59.9 \\ 
    add Attention &52.2 &67.5 &63.7 &57.8 &\textbf{75.8} &57.9 &57.9 &57.9 \\
    add DMFM &\textbf{54.3} &\textbf{68.9} &\textbf{65.0} &\textbf{58.3} &75.7 &\textbf{59.9} &\textbf{59.4} &\textbf{60.5}\\
    \bottomrule 
    \end{tabular}
    }
\end{table}

    \textbf{Analysis of the STMA}. Spatio-Temporal Mamba Adapter (STMA) unifies cross-modal data interaction and spatio-temporal cue modeling.
    We design experiments to prove STMA's effectiveness and discuss the relationship between STMA's number of layers and efficiency.

    To verify STMA’s effectiveness, we conducted experiments in Tab.\ref{tab:ablationtabs5}. Starting from the baseline, we develop four cross-modal interaction variants: (\textrm{I}) Attention-based, (\textrm{II}) Mamba-based, (\textrm{III}) Mamba with MLP, and (\textrm{IV}) Mamba with MultiFFT. Comparing \textrm{I} and \textrm{IV} reveals that \textrm{IV} achieves higher performance with fewer parameters, highlighting STMA’s advantage over attention. To assess MultiFFT’s effect, we compared \textrm{II} vs. \textrm{IV} and \textrm{III} vs. \textrm{IV}. These comparisons confirm that MultiFFT greatly enhances Mamba’s tracking performance while adding minimal parameters.

    To analyze the impact of STMA layer count on efficiency, we conduct experiments shown in Tab.\ref{tab:ablationtabs6}. We create six {\tracker} variants, each with a different number of STMA layers. Results show that increasing STMA layers improves performance, but also adds parameters and slows down speed. Optimal performance and a balance between speed and accuracy are achieved at six layers.

    \textbf{Analysis of DMFM}. To demonstrate the effectiveness of our Dynamic Multi-modal Feature Mixer (DMFM) for final multi-modal feature fusion in tracking, we conduct experiments based on the Baseline in Tab.\ref{tab:ablationtabs7}. We design four variants using convolution, MLP, attention and DMFM to explore different fusion methods. Comparing results across these four variants in three modalities, DMFM shows the best performance. Compared to the other three methods, DMFM consistently performs best across modalities, proving to be a more stable and effective fusion method for multi-modal features.

\subsection{Visualization and Speed Analysis}

\textbf{Visualization Analysis}.
To demonstrate the effectiveness of the proposed multi-modal tracking algorithm {\tracker} in real-world scenarios, we conduct comprehensive visualization analyses of tracking tasks involving infrared, depth, and event modalities. We compare {\tracker} with three state-of-the-art RGB-T trackers on six sequences from the LasHeR \cite{LasHeR}, VisEvent \cite{VisEvent}, and DepthTrack \cite{DeT} datasets, as shown in Fig.\ref{fig:drawtrackers}. Benefiting from the proposed spatio-temporal mamba adapter and dynamic multi-modal feature mixer, the proposed {\tracker} demonstrates superior cross-modal complementary perception compared with other RGB-E, RGB-D, and RGB-T trackers UnTrack \cite{un-track}, SDSTrack \cite{SDSTrack}, and ViPT \cite{ViPT}. These results highlight {\tracker}’s superior target identification capabilities. Additionally, Fig.\ref{fig:attributes} shows {\tracker} consistently outperforms other trackers across all attributes on LasHeR \cite{LasHeR} and VisEvent \cite{VisEvent}, confirming its effectiveness and adaptability under various challenging attributes. Finally, in Fig.\ref{fig:drawattn}, we also visualize attention maps on these three dataset, comparing the baseline (in Tab.\ref{tab:ablationtabs4}) and {\tracker}. The comparison reveals {\tracker}’s ability to focus attention more precisely on target objects, demonstrating stronger discriminative capabilities.

\textbf{Speed Analysis}. We comprehensively evaluate the efficiency of our model on an NVIDIA V100 GPU, considering module parameters, module computational complexity, and model inference speed to verify its advantages in modeling complexity. As shown in Tab.\ref{tab:speed_compare}, SDSTrack adopts a symmetric multi-modal adapter with 9.47M parameters to integrate visible, infrared, depth, or event features. In comparison, our {\tracker} achieves a significant reduction in the learnable parameters. Specifically, the proposed spatio-temporal mamba adapter and dynamic multi-modal feature mixer are both simple and effective, containing only 11.9M parameters, yielding a 2.9M reduction compared to SDSTrack. Benefiting from these advantages, UBATrack-256 achieves 32.5 FPS in multi-modal tracking.

\section{Conclusion}
\label{sec:conlusion}

This paper has proposed a unified multi-modal tracking framework, {\tracker}, which is designed to efficiently capture spatio-temporal dependencies across modalities. Inspired by Mamba's linear complexity and long-sequence modeling, we have introduced the Spatio-Temporal Mamba Adapter (STMA), which unifies cross-modal interaction and spatio-temporal modeling. Additionally, we have proposed a Dynamic Multi-modal Feature Mixer that has focused more on high-level semantic information during final feature fusion for more effective integration. Experiments on various modal tracking benchmarks have shown that {\tracker} has achieved state-of-the-art performance.

\section*{Acknowledgments}

This work is supported by the Project of Guangxi Science and Technology (No. 2025GXNSFAA069676, 2025GXNSFAA069417, 2024GXNSFGA010001, and GuiKeFN2504240017), the National Natural Science Foundation of China (No.U23A20383, 62472109 and 62466051), the Guangxi ”Young Bagui Scholar” Teams for Innovation and Research Project, the Research Project of Guangxi Normal University (No. 2025DF001).


\bibliographystyle{IEEEtran}
\bibliography{main}


\end{document}